\documentclass[10pt]{article}
\usepackage{arxiv}
\usepackage{cite}
\usepackage{amsmath,amssymb,amsfonts}
\usepackage{algorithmic}
\usepackage{graphicx}
\usepackage{textcomp}
\usepackage{multirow}
\usepackage{booktabs}
\usepackage{url}
\usepackage{bm}
\usepackage[colorlinks=true,linkcolor=blue,citecolor=blue,urlcolor=blue]{hyperref}

\newcommand{\orcidicon}[1]{\href{https://orcid.org/#1}{\includegraphics[width=0.32cm]{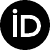}}}

\def\BibTeX{{\rm B\kern-.05em{\sc i\kern-.025em b}\kern-.08em
    T\kern-.1667em\lower.7ex\hbox{E}\kern-.125emX}}

\title{Cosine-Similarity Routing with Semantic Anchors\\ for Interpretable Mixture-of-Experts Language Models}

\author{
  Ivan Ternovtsii \orcidicon{0009-0009-9267-8516}\thanks{This research was conducted as part of PhD studies at the Department of Software Systems, Faculty of Information Technologies, Uzhhorod National University. HengeBytes generously provided computational resources. We thank the Department of Software Systems for academic support and the HengeBytes team for maintaining the computational infrastructure. Corresponding author: Ivan Ternovtsii (e-mail: ivan.ternovtsii@uzhnu.edu.ua).} \\
  Department of Software Systems, Uzhhorod National University\\
  Narodna sq. 3, Uzhhorod, Ukraine, 88000\\
  HengeBytes\\
  \texttt{ivan.ternovtsii@uzhnu.edu.ua} \\
  \And
  Yurii Bilak \orcidicon{0000-0001-5989-1643} \\
  Department of Software Systems, Uzhhorod National University\\
  Narodna sq. 3, Uzhhorod, Ukraine, 88000\\
}

\date{March 29, 2026}

\begin{document}

\maketitle

\begin{abstract}
Mixture-of-Experts (MoE) models improve efficiency through sparse activation, but their learned gating functions provide limited insight into routing decisions. This work introduces the Semantic Resonance Architecture (SRA), which routes tokens to experts via cosine similarity between token representations and learnable semantic anchors, making every routing decision directly traceable to anchor--token similarity scores. We evaluate SRA on WikiText-103 across 17 configurations. In a controlled multi-seed comparison (3 seeds $\times$ 4 configurations, 256 experts, $D_{ff}$=256), cosine routing achieves competitive perplexity with standard linear routing (12.57$\pm$0.03 vs 12.45$\pm$0.03 for K=1$\to$4; 12.52$\pm$0.02 vs 12.57$\pm$0.02 for K=2$\to$4). The training recipe---not the routing function---drives specialization quality, while cosine routing provides inherent inspectability. We introduce a bandpass routing loss---a floor-and-ceiling corridor on expert utilization---that reduces dead experts from 30--32\% (under earlier configurations) to 0--6\% and transfers to both routing types. Routing-space evaluation shows cosine routing provides significantly better word-level subtoken coherence in deeper layers ($p<$0.001), with 44--54\% of expert specialization being syntactic rather than semantic. Extended analysis reveals cosine routing maintains more stable router saturation and tighter per-expert vocabulary distributions---structural advantages from the bounded cosine similarity range. An inference-time $k$-sweep shows that $k$=5 yields a free $\sim$0.1--0.2 perplexity gain over the best $k$=4 validation PPL. Cross-dataset validation on OpenWebText confirms generalization: cosine routing achieves comparable perplexity (44.88 vs 45.44), the bandpass loss eliminates dead experts, and specialization patterns are preserved.
\end{abstract}


\section{Introduction}
\label{sec:introduction}
The rapid advancement of large language models (LLMs) has transformed natural language processing \cite{Brown2020, Chowdhery2023}. As these models are deployed in increasingly sensitive domains, understanding their internal decision-making processes becomes correspondingly important \cite{Rudin2019}. Interpretability exists on a spectrum, ranging from post-hoc analysis of model outputs to architectures that provide built-in transparency at specific levels of abstraction.

Mixture-of-Experts (MoE) architectures address the computational challenges of scaling by conditionally activating only a subset of parameters per token \cite{Shazeer2017}. However, existing routing mechanisms in models such as the Switch Transformer \cite{Fedus2022} and GShard \cite{Lepikhin2021} rely on learned linear gating networks whose internal logic is opaque. A practitioner can observe \emph{which} expert was selected, but the \emph{reason} for that selection requires post-hoc investigation of the gate weights.

This work develops the Semantic Resonance Architecture (SRA), which replaces learned linear gating with cosine-similarity routing through learnable semantic anchors. In SRA, the Chamber of Semantic Resonance (CSR) module computes normalized dot products between token representations and per-expert anchor vectors, making the routing score an explicit measure of semantic similarity. This provides interpretability at the routing level: every routing decision can be directly explained by examining which anchors a token is most similar to.

Our contributions are threefold. First, we introduce cosine-similarity routing with semantic anchors (Section~\ref{sec:sra}), providing an inspectable routing mechanism where decisions are grounded in measurable similarity scores. Second, we propose the bandpass routing loss (Section~\ref{sec:bandpass}), a floor-and-ceiling corridor constraint that substantially improves expert utilization and transfers effectively to both cosine and linear routing. Third, we present comprehensive ablations across 17 configurations (Section~\ref{sec:experiment}), including quantitative interpretability metrics (Section~\ref{sec:specialization}) and extended specialization analysis (Section~\ref{sec:extended_metrics}), that reveal the training recipe---not the routing function---as the primary driver of semantic specialization, while cosine routing provides inherent inspectability by design. Robustness analysis (Section~\ref{sec:robustness}) and deployment considerations (Section~\ref{sec:deployment}) further validate the practical utility of the architecture.

\section{Research Goals and Objectives}
\label{sec:goals}
The goal of this research is to develop and evaluate the Semantic Resonance Architecture (SRA), which provides an inspectable routing mechanism for Mixture-of-Experts models based on cosine similarity.

To achieve this goal, the following objectives were defined:
\begin{itemize}
    \item Analyze existing routing methods in MoE models and outline their key shortcomings related to opacity and instability.
    \item Develop cosine-similarity routing as a mechanism for inspectable expert selection based on semantic similarity.
    \item Design and implement the Chamber of Semantic Resonance (CSR), which connects token representations with semantic anchors.
    \item Experimentally evaluate SRA compared to traditional dense and MoE architectures on the WikiText-103 corpus with comprehensive ablations.
    \item Conduct a quantitative interpretability analysis of expert behavior using Internal Cohesion and External Purity metrics.
    \item Discuss potential applications of the architecture in settings where routing-level interpretability is valuable.
\end{itemize}

\section{Related Work}
\subsection{Mixture-of-Experts Routing}
The Mixture-of-Experts paradigm was proposed by Jacobs \textit{et al.} \cite{Jacobs1991} and Jordan and Jacobs \cite{Jordan1994}. Shazeer \textit{et al.} \cite{Shazeer2017} introduced the Sparsely-Gated MoE layer for transformers, demonstrating that conditional computation can increase model capacity without proportional cost increase. The Switch Transformer \cite{Fedus2022} simplified routing to top-1, achieving significant speedups. GShard \cite{Lepikhin2021} scaled MoE to multilingual tasks. These methods use learned linear gates followed by softmax, which offer limited semantic interpretation of routing decisions.

Beyond learned linear routing, alternative approaches have been explored. Hash-based routing \cite{Roller2021} assigns tokens to experts via deterministic hash functions, eliminating learnable routing parameters entirely but sacrificing adaptivity. Expert Choice routing \cite{Zhou2022} inverts the assignment direction---experts select tokens rather than tokens selecting experts---guaranteeing perfect load balance by construction, though it is incompatible with autoregressive generation. Soft MoE \cite{Puigcerver2024} eliminates discrete routing entirely by passing weighted token mixtures to all experts. DEMix \cite{Gururangan2022} routes by domain, and BASE layers \cite{Lewis2021} use linear assignment for balanced routing. ReLU-based routing \cite{Wang2025} replaces top-$k$ selection with ReLU gating, enabling variable expert counts per token.

Most closely related to our approach, Chi \textit{et al.} \cite{Chi2022} proposed X-MoE, which routes tokens on a low-dimensional hypersphere using L2 normalization and cosine similarity with a gating temperature. Our work extends this geometric routing framework with three key additions: (1) a bandpass routing loss that provides explicit corridor-based control over expert utilization, (2) quantitative interpretability metrics (IC/EP) evaluated via an independent sentence-transformer judge, and (3) a matched routing comparison demonstrating that cosine routing's advantage over linear routing lies in inherent inspectability rather than improved specialization quality. Nguyen \textit{et al.} \cite{Nguyen2025} subsequently provided theoretical analysis of cosine routers, showing that perturbation improves convergence rates.

Recent large-scale MoE architectures provide context for our design choices. DeepSeekMoE \cite{Dai2024} demonstrated that fine-grained expert segmentation (many small experts) improves specialization, consistent with our 256-expert design. DeepSeek-V3 \cite{DeepSeekV3} introduced auxiliary-loss-free load balancing via dynamic per-expert bias terms, representing an alternative to our bandpass loss: their approach eliminates auxiliary loss interference but provides less explicit control over utilization corridors.

Analysis of expert specialization in MoE models has received increasing attention. OLMoE \cite{Muennighoff2024} defined four analysis metrics (router saturation, expert co-activation, domain specialization, vocabulary specialization) for a fully open 1B/7B MoE model. OpenMoE \cite{Xue2024} identified three routing phenomena---context-independent specialization, early routing learning, and drop-towards-the-end---suggesting that routing assignments crystallize early in training. Mixtral \cite{Jiang2024} found minimal domain-level specialization despite strong performance, supporting our finding that routing quality depends more on the training recipe than the routing mechanism.

\subsection{Interpretability Approaches}
Research in neural network interpretability spans multiple levels of analysis. Attention visualization \cite{Vaswani2017, Clark2019} reveals which input positions influence outputs. Probing classifiers \cite{Tenney2019, Hewitt2019} test whether specific linguistic features are encoded in hidden representations. Mechanistic interpretability\footnote{N. Elhage \emph{et al.}, ``A mathematical framework for transformer circuits,'' Transformer Circuits Thread, 2021.}\textsuperscript{,}\footnote{C. Olsson \emph{et al.}, ``In-context learning and induction heads,'' Transformer Circuits Thread, 2022.} aims to reverse-engineer the computational circuits within neural networks.

It is important to distinguish routing-level interpretability from mechanistic interpretability. Mechanistic methods aim to understand the full computational graph---how individual neurons, attention heads, and circuits produce outputs. Routing-level interpretability, as provided by SRA, reveals \emph{which expert processes which token and why} (via similarity scores), without claiming to explain the expert's internal computation. This distinction is practically significant: routing interpretability is available even for restricted-access or commercial models where internal weights cannot be inspected, whereas mechanistic interpretability requires full model access.

Model-agnostic explainability methods such as LIME \cite{Ribeiro2016}, SHAP \cite{Lundberg2017}, and toolkit-based approaches including IBM's AI Explainability 360 \cite{Arora2022} and gSMILE \cite{Jia2024} explain individual predictions by perturbing inputs, computing feature attributions, or generating local surrogate models. These methods operate at the input--output level and are complementary to SRA's routing-level interpretability: model-agnostic methods explain \emph{what} the model predicts, while SRA reveals \emph{how} the model internally routes information through its expert network. The two layers of explanation can be combined---for instance, one could use SHAP to identify which input tokens most influence an output, and then examine SRA's routing to understand which experts processed those critical tokens.

\subsection{Semantic Representations}
The use of semantic embeddings capturing relationships through distributional semantics \cite{Mikolov2013, Pennington2014} is foundational to modern NLP. Pre-trained language models learn rich, contextualized representations \cite{Peters2018, Devlin2019}. This work leverages these insights by using cosine similarity in the learned representation space as the basis for routing decisions.

The development of accurate and interpretable models extends beyond NLP into applied sciences. Research in computational physics and spectroscopy has shown significant progress through machine learning methods \cite{Bilak2025a, Bilak2025b, Bilak2025c}, demonstrating the broad need for models that combine accuracy with interpretability across scientific domains.

\section{Methods and Models}
\label{sec:sra}

\subsection{Architecture Overview}
The Semantic Resonance Architecture is based on the standard transformer decoder, replacing the feed-forward network (FFN) at each layer with the Chamber of Semantic Resonance (CSR) module. The architecture uses learned embeddings with weight tying to the output projection and Rotary Position Embeddings (RoPE) \cite{Su2024} for relative position information, with a Pre-LayerNorm configuration.

For an input sequence $X \in \mathbb{R}^{B \times L \times D}$, where $B$ is the batch size, $L$ is the sequence length, and $D$ is the model dimension, each SRA block computes:
\begin{align*}
X' &= \text{LayerNorm}(X) \\
H &= X + \text{MultiHeadAttention}(X') \\
H' &= \text{LayerNorm}(H) \\
Y &= H + \text{CSR}(H')
\end{align*}

\subsection{Chamber of Semantic Resonance}
The CSR module routes tokens based on their similarity with learnable semantic anchors, making routing decisions directly traceable.

\subsubsection{Semantic Anchors and Initialization}
We initialize a set of learnable semantic anchors $A \in \mathbb{R}^{N \times D}$, where $N$ is the number of experts. Using orthogonal initialization to maximize initial dispersion:
\[ A = \text{orth}(\text{randn}(N, D)) \]
We also evaluate batch seeding---initializing anchors from real token embeddings on the first forward pass---which captures the Zipfian distribution of tokens and reduces early dead experts.

\subsubsection{Resonance Calculation}
For each token representation $h \in \mathbb{R}^{D}$, we compute resonance scores using cosine similarity, scaled by a learnable global temperature parameter $\tau$ (initialized to 10.0):
\begin{equation}
r_{i} = \tau \cdot \cos(h, a_{i}) = \tau \cdot \frac{h \cdot a_{i}}{\|h\|_{2} \cdot \|a_{i}\|_{2} + \epsilon}
\label{eq:resonance}
\end{equation}
where $a_{i}$ is the $i$-th semantic anchor, $\epsilon = 10^{-8}$, and the computation is performed in FP32 for numerical stability during mixed-precision training. The temperature $\tau$ controls routing sharpness: higher values produce more peaked distributions, while lower values yield more uniform routing.

\subsubsection{Top-k Expert Selection and Execution}
We compute softmax routing probabilities over \emph{all} $N$ experts before top-$k$ selection, ensuring gradient flow to all anchors even when $k=1$:
\begin{align}
p &= \text{softmax}(r) \label{eq:full_softmax} \\
\text{indices} &= \text{top\_k\_indices}(p, k) \nonumber \\
w_{i} &= \frac{p_{i}}{\sum_{j \in \text{indices}} p_{j}}, \quad i \in \text{indices} \nonumber
\end{align}
The final output is a weighted combination of the selected experts' outputs:
\begin{equation}
y = \sum_{i \in \text{indices}} w_{i} \cdot \text{Expert}_{i}(h)
\label{eq:expert_output}
\end{equation}
Each expert is a two-layer feed-forward network with GELU activation.

\subsection{Training Objectives}
The total loss function combines the primary language modeling objective with auxiliary load balancing:
\begin{equation}
\mathcal{L} = \mathcal{L}_{LM} + \alpha \cdot \mathcal{L}_{balance}
\label{eq:total_loss}
\end{equation}

\subsubsection{Bandpass Routing Loss}
\label{sec:bandpass}
We introduce the bandpass routing loss, which constrains per-expert token fractions within a floor-and-ceiling corridor. Let $f_i$ denote the mean routing probability for expert $i$ (the softmax output averaged over all tokens in the batch). The bandpass loss penalizes experts that fall below a minimum utilization floor or exceed a maximum utilization ceiling:
\begin{equation}
\mathcal{L}_{balance} = \frac{1}{N} \sum_{i=1}^{N} \left[ \max(0, f_{min} - f_i)^2 + \max(0, f_i - f_{max})^2 \right]
\label{eq:bandpass_loss}
\end{equation}
where $f_{min}$ and $f_{max}$ define the acceptable corridor, scaled relative to the uniform rate $1/N$. Concretely, with $N$=256 experts and the uniform rate $u=1/N$, we set $f_{\min}=0.05\%$ and $f_{\max}=0.40\%$ (i.e., $\approx$0.13$u$ and $\approx$1.02$u$). The ceiling (anti-monopoly) parameter is the dominant factor: setting $f_{max}$ near the uniform rate forces aggressive redistribution that monotonically improves perplexity across corridor configurations. The total loss is $\mathcal{L} = \mathcal{L}_{LM} + \alpha \cdot \mathcal{L}_{balance}$ with $\alpha$=0.4.

\textbf{PAD masking.} When computing $f_i$, we exclude padding tokens (which constitute $\sim$24\% of batched sequences) from both the token fractions and the loss computation. Without masking, the router wastes expert capacity on uninformative padding tokens, contaminating the routing statistics and inflating dead expert counts.

As a baseline, we also evaluate the squared coefficient of variation (CV$^2$) loss \cite{Fedus2022}:
\begin{equation}
\mathcal{L}_{CV^2} = N \cdot \frac{\text{Var}(P_{mean})}{\text{Mean}(P_{mean})^{2} + \epsilon}
\label{eq:cv_loss}
\end{equation}
where $P_{mean}$ is the mean routing probability across the batch.

\subsection{Progressive Routing}
Starting training with top-$k$>1 often leads to expert collapse, where many experts remain permanently unused. We employ a progressive routing strategy: training begins with top-1 routing, allowing each expert to establish distinct specialization, then transitions to top-4 at epoch 3 (the ``BandAid'' schedule). This K=1$\rightarrow$4 schedule outperforms both K=1$\rightarrow$2 and the three-step K=1$\rightarrow$2$\rightarrow$4 schedule.

We also evaluate K=2$\rightarrow$4 warmup (starting directly at top-2), which avoids the top-1 information bottleneck and produces zero dead experts at 256-expert scale, at the cost of slightly reduced interpretability metrics.

\section{Experiment}
\label{sec:experiment}
The WikiText-103 corpus \cite{Merity2017} was used as the primary evaluation benchmark, comprising 103 million tokens from Wikipedia articles with byte-pair encoding (vocabulary size 32,000). Cross-dataset validation on OpenWebText \cite{Gokaslan2019} is presented in Section~\ref{sec:cross_dataset}.

\subsection{Model Configurations}
Three model families were compared, all with $D$=512, four layers, and eight attention heads:

\textbf{SRA (best configuration):} 256 experts per layer (1,024 total), $D_{ff}$=512, progressive K=1$\rightarrow$4 at epoch 3. Total parameters: $\approx$559M, active parameters: $\approx$29M. Balance coefficient $\alpha$=0.4, temperature $\tau$=10.0 (learnable).

\textbf{Standard MoE:} 128 experts per layer, $D_{ff}$=1024, progressive K=1$\rightarrow$2 at epoch 6, learned linear router. Total parameters: $\approx$559M, active parameters: $\approx$29M.

\textbf{Dense baseline:} $D_{ff}$=2048. Total parameters: $\approx$29M, active parameters: $\approx$29M.

The 559M models (Table~\ref{tab:main_results}, rows 2--4) were trained for 10 epochs; the matched comparison and K-schedule ablation (Tables~\ref{tab:routing_comparison}--\ref{tab:k_schedule}) used 8 epochs. All models used AdamW ($\beta_1$=0.9, $\beta_2$=0.95), learning rate $3 \times 10^{-4}$ with linear warmup over 4,000 steps and cosine decay, batch size 128, sequence length 256, dropout 0.1, and weight decay 0.01 (with temperature excluded from weight decay). Gradient checkpointing was enabled for the 256-expert configuration to reduce VRAM usage ($\sim$60\% savings at $\sim$15\% compute overhead). Experiments were conducted on NVIDIA RTX 3090 GPUs.

\subsection{Matched Routing Comparison}
To isolate the effect of the routing function, we trained matched cosine and linear routing models with identical configurations: 256 experts, $D_{ff}$=256, K=1$\rightarrow$4 at epoch 3, bandpass loss ($\alpha$=0.4), and PAD masking. Both K=1$\rightarrow$4 (B schedule, transition at epoch 3) and K=2$\rightarrow$4 (E schedule, transition at epoch 3) were tested.

\section{Results}

We organize results by three experimental dimensions. Table~\ref{tab:main_results} compares configurations at varying scales (single seed unless noted). Table~\ref{tab:routing_comparison} provides a controlled routing-mechanism comparison at fixed scale with multiple seeds---this is the primary evidence for evaluating the cosine routing contribution. Table~\ref{tab:k_schedule} ablates the progressive $k$-schedule. Readers should compare within tables, not across them, as model sizes differ.

\subsection{Main Results}
Table~\ref{tab:main_results} presents perplexity across configurations at varying scales.

\begin{table}[h]
\caption{\textbf{Perplexity comparison on WikiText-103 validation set (single seed unless noted). Active parameters ($\approx$29M) are matched for top four models.}}
\label{tab:main_results}
\centering
\begin{tabular}{|l|c|c|c|c|}
\hline
Model & Config & Total & Val PPL & Dead Exp \\
\hline
Dense Baseline & $D_{ff}$=2048 & 29M & 14.13 & N/A \\
Standard MoE & 128$\times$1024, K=1$\to$2 & 559M & 12.76 & 165 (32\%) \\
SRA (128 exp) & 128$\times$1024, K=1$\to$2 & 559M & 12.72 & 44 (9\%) \\
SRA (256 exp)$^{\dagger}$ & 256$\times$512, K=1$\to$4 & 559M & 12.20 & 305 (30\%)$^{*}$ \\
SRA-B (256 exp) & 256$\times$256, K=1$\to$4 & 295M & 12.57$\pm$0.03 & 0--6\% \\
SRA-E (256, opt.) & 256$\times$256, K=2$\to$4 & 295M & \textbf{12.52$\pm$0.02} & \textbf{0 (0\%)} \\
\hline
\multicolumn{5}{l}{\footnotesize $^{*}$Early bandpass config; optimized corridor reduces dead to 0--6\% (Table~\ref{tab:k_schedule}).} \\
\multicolumn{5}{l}{\footnotesize $^{\dagger}$Single seed (seed=19). Multi-seed validation not performed at this scale.}
\end{tabular}
\end{table}

The primary result is the multi-seed validated SRA-E configuration: 12.52$\pm$0.02 PPL across 3 seeds with zero dead experts and half the total parameters of the 559M models. The controlled multi-seed comparison in Section~\ref{sec:routing_comparison} provides the primary evidence for evaluating the routing mechanism. At the matched 128-expert configuration, SRA (12.72) and the standard MoE (12.76) achieve comparable perplexity, but SRA exhibits fewer dead experts (9\% vs 32\%). The K=2$\to$4 schedule eliminates dead experts entirely for both routing types, demonstrating that the training recipe can be tuned to prioritize either raw perplexity or expert health. A larger single-seed run (256$\times$512, 559M params) reached 12.20 PPL but used an early bandpass corridor with 30\% dead experts; this suggests further gains from scaling expert capacity, though it requires multi-seed replication with the optimized corridor. The SRA-B configuration (12.57$\pm$0.03 across 3 seeds) provides the matched-pair comparison point for routing type analysis.

\subsection{Routing Type Comparison}
\label{sec:routing_comparison}
To isolate the effect of the routing function from the training recipe, we trained matched cosine and linear routing models with identical configurations (Table~\ref{tab:routing_comparison}).

\begin{table}[h]
\caption{\textbf{Matched routing comparison (256 experts, $D_{ff}$=256, K=1$\to$4, bandpass loss). Val PPL: mean$\pm$std across 3 seeds (19, 42, 137). Test PPL: seed=19.}}
\label{tab:routing_comparison}
\centering
\begin{tabular}{|l|c|c|c|c|c|}
\hline
Routing & Val PPL & Test PPL & Dead & IC & EP \\
\hline
Linear (StdMoE B) & \textbf{12.45$\pm$0.03} & 13.01 & 0--5\% & 0.250$\pm$0.011 & 0.361$\pm$0.009 \\
Cosine (SRA B) & 12.57$\pm$0.03 & \textbf{12.96} & 0--6\% & 0.232$\pm$0.004 & 0.362$\pm$0.001 \\
\hline
\end{tabular}
\end{table}

With the same training recipe across 3 random seeds, linear routing achieves slightly better perplexity (12.45$\pm$0.03 vs 12.57$\pm$0.03) and comparable interpretability metrics. The 0.12 PPL gap is consistent across seeds, confirming it is a real but small effect rather than seed variance. On the test set (seed=19), the gap narrows: SRA~B achieves 12.96 vs StdMoE~B at 13.01, suggesting comparable generalization. This result indicates that, on this task, semantic specialization is primarily driven by the training recipe---bandpass loss, progressive routing, and PAD masking---rather than by the routing function itself. The advantage of cosine routing lies not in emergent specialization quality but in \emph{inherent inspectability}: routing decisions are grounded in cosine similarity between tokens and interpretable anchor vectors, which can be examined directly without post-hoc analysis of opaque gate weights.

\subsection{Training Dynamics}
\begin{figure}[t]
\centering
\includegraphics[width=\textwidth]{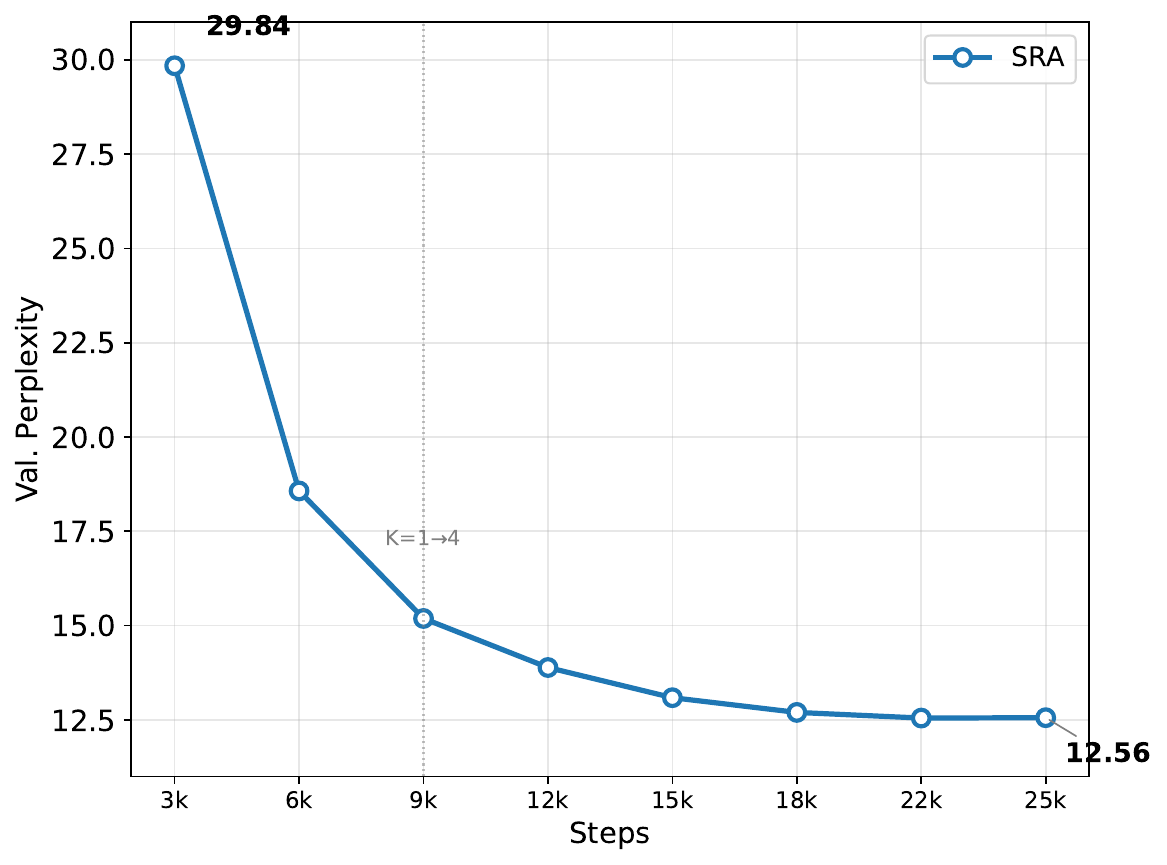}
\caption{\textbf{SRA-B validation perplexity (256$\times$256, K=1$\to$4, seed=42, 295M).}\label{fig:sra_perplexity}}
\end{figure}

Figure~\ref{fig:sra_perplexity} shows the SRA validation perplexity curve during training.

The SRA model shows a characteristic pattern: during the top-1 phase (epochs 1--2), specialization develops but performance lags the dense baseline (Table~\ref{tab:main_results}). The transition to top-4 at epoch 3 produces a clear inflection point as the model begins leveraging expert combinations. The same inflection pattern is observed across all configurations, including the 295M multi-seed runs (Section~\ref{sec:routing_comparison}).

\subsection{Bandpass Loss Ablation}
Table~\ref{tab:bandpass_ablation} compares auxiliary loss functions, all with PAD masking enabled (128 experts, $D_{ff}$=256, K=1$\rightarrow$2 at epoch 3).

\begin{table}[h]
\caption{\textbf{Auxiliary loss comparison with PAD masking (128 experts, $D_{ff}$=256).}}
\label{tab:bandpass_ablation}
\centering
\begin{tabular}{|l|c|c|c|}
\hline
Loss Type & Corridor & Val PPL & Dead Exp \\
\hline
CV$^2$ (masked) & --- & 13.28 & 46 (9\%) \\
Elastic Ceiling CV$^2$ & ceil=0.010 & 13.08 & 30 (6\%) \\
\textbf{Bandpass S6 (masked)} & \textbf{0.003--0.008} & \textbf{13.04} & \textbf{29 (6\%)} \\
\hline
Bandpass S6 (unmasked) & 0.003--0.008 & 13.19 & 96 (19\%) \\
CV$^2$ (unmasked) & --- & 13.40 & 59 (12\%) \\
\hline
\end{tabular}
\end{table}

Two findings emerge. First, bandpass loss with a near-uniform ceiling (102\% of $1/N$) outperforms CV$^2$ on both perplexity and expert health. The ceiling parameter is dominant: perplexity improves monotonically as the ceiling tightens toward the uniform rate. Second, PAD masking is a substantial improvement across all loss types---it reduces dead experts from 19\% to 6\% for bandpass and prevents the router from wasting capacity on uninformative padding tokens.

\subsection{K-Schedule Ablation}
\label{sec:k_schedule}
Table~\ref{tab:k_schedule} presents results from seven configurations varying expert count (128 vs 256) and progressive $k$ schedule, all using masked bandpass loss.

\begin{table}[h]
\caption{\textbf{K-schedule ablation with bandpass loss (all models: cosine routing, $D_{ff}$=256, masked bandpass, 8 epochs). PPL for Runs B and E is mean$\pm$std across 3 seeds; others are single seed (seed=19). IC/EP via independent sentence-transformer judge (seed=19).}}
\label{tab:k_schedule}
\centering
\begin{tabular}{|l|c|c|c|c|c|c|}
\hline
Run & $N_{exp}$ & K Schedule & PPL & Dead & IC & EP \\
\hline
E & 256 & K=2$\to$4 & \textbf{12.52$\pm$0.02} & \textbf{0\%} & 0.221 & 0.328 \\
B & 256 & K=1$\to$4 & 12.57$\pm$0.03 & 0--6\% & 0.227 & 0.363 \\
D & 256 & K=1$\to$2$\to$4 & 12.66 & 1\% & 0.217 & 0.357 \\
F & 128 & K=2$\to$4 & 12.67 & 0\% & 0.208 & 0.329 \\
A & 128 & K=1$\to$4 & 12.75 & 0\% & 0.240 & 0.362 \\
G & 128 & K=1$\to$2$\to$4 & 12.84 & 0\% & 0.238 & 0.361 \\
C & 256 & K=1$\to$2 & 12.88 & 16\% & 0.223 & 0.358 \\
\hline
\end{tabular}
\end{table}

Three patterns emerge: (1) \textbf{Compositionality matters more than capacity.} K=4 with 128 experts (Run A, 12.75) outperforms K=2 with 256 experts (Run C, 12.88). (2) \textbf{K=2 warmup eliminates dead experts.} Run E (K=2$\rightarrow$4) achieves 0\% dead experts at 256-expert scale, compared to 0--6\% for K=1$\rightarrow$4. (3) \textbf{256 experts consistently outperform 128} at the same $k$-schedule (Run E 12.52 vs Run F 12.67, Run B 12.57 vs Run A 12.75).

\subsection{Inference-Time K-Sweep}
\label{sec:k_sweep}
At inference time, $k$ can be set freely regardless of training schedule. Table~\ref{tab:k_sweep} shows that K=5 is optimal or tied-optimal across all models trained with K$\rightarrow$4, providing a free $\sim$0.1--0.2 perplexity improvement over the best K=4 validation PPL from training.

\begin{table}[h]
\caption{\textbf{Inference-time K-sweep (256-expert cosine models, seed=19, final checkpoint). Bold = best PPL.} K=4 values may differ from Table~\ref{tab:k_schedule} because that table reports best validation PPL during training.}
\label{tab:k_sweep}
\centering
\begin{tabular}{|l|c|c|c|c|c|c|}
\hline
Model (Train K) & K=2 & K=4 & \textbf{K=5} & K=6 & K=8 & K=12 \\
\hline
K=2$\to$4 (E) & 13.93 & 12.61 & \textbf{12.36} & 12.43 & 12.63 & 13.41 \\
K=1$\to$4 (B) & 25.48 & 12.97 & \textbf{12.39} & 12.42 & 12.61 & 13.16 \\
K=1$\to$2$\to$4 (D) & 22.43 & 13.31 & \textbf{12.56} & \textbf{12.56} & 12.72 & 13.11 \\
K=1$\to$2 (C) & 22.94 & 14.24 & \textbf{13.07} & 13.29 & 13.82 & 15.21 \\
\hline
\end{tabular}
\end{table}

This result suggests that models learn slightly more expert combinations than they utilize during training, and one additional expert at inference can capture these latent combinations. Models trained with higher $k$ degrade more gracefully across inference $k$ values, while K=1$\rightarrow$2 models show rapid degradation above K=5.

\subsection{Expert Specialization}
\label{sec:specialization}
Analysis of expert specialization reveals that SRA experts develop distinct semantic and syntactic roles. Table~\ref{tab:sra_specialization} shows examples from Layer 1, where experts form stable correspondences with specific linguistic categories.

\begin{table}[h]
\caption{\textbf{Expert specialization in SRA (Layer 1, 256 experts, 559M model). Experts learn coherent semantic/syntactic categories via anchor--token similarity.}}
\label{tab:sra_specialization}
\centering
\begin{tabular}{|l|l|l|}
\hline
Expert & Category & Top Tokens \\
\hline
$E_{6}$ & Weather systems & tropical, Hurricane, storm, cyclone \\
$E_{138}$ & Superlatives & most, best, largest, highest, greatest \\
$E_{191}$ & Possessive pron. & their, its, own, your \\
$E_{80}$ & Temporal & early, late, April, March, September \\
$E_{157}$ & Education & school, football, students, college \\
$E_{205}$ & Causative verbs & made, making, make, give, gave \\
\hline
\end{tabular}
\end{table}

The specialization pattern follows a hierarchical organization across layers: Layer 0 captures orthographic/subword patterns, Layer 1 develops thematic clusters (weather, education, legal, temporal) alongside grammatical specialists (possessives, superlatives, causative verbs), Layer 2 handles structural/transitional patterns, and Layer 3 captures abstract semantic and tense distinctions.

\subsubsection{Quantitative Interpretability Metrics}
We define two metrics evaluated by an independent sentence-transformer model (all-MiniLM-L6-v2):

\textbf{Internal Cohesion (IC):} The average pairwise cosine similarity of an expert's top-20 routed tokens in the external embedding space. Higher IC indicates the expert processes tokens that are semantically related.

\textbf{External Purity (EP):} The average pairwise cosine similarity of an expert's top-20 tokens as judged by the independent model. Higher EP indicates the expert's tokens form a semantically coherent cluster from an external perspective.

Table~\ref{tab:ic_ep} presents IC and EP across representative configurations.

\begin{table}[h]
\caption{\textbf{Interpretability metrics across configurations (routing-based IC/EP). All models: 256 experts, $D_{ff}$=256, bandpass loss. Val PPL: mean$\pm$std across 3 seeds (19, 42, 137). Test PPL: seed=19.}}
\label{tab:ic_ep}
\centering
\begin{tabular}{|l|c|c|c|c|}
\hline
Model (Routing) & Val PPL & Test PPL & IC & EP \\
\hline
StdMoE B, K=1$\to$4 (Lin.) & \textbf{12.45$\pm$0.03} & 13.01 & 0.250$\pm$0.011 & 0.361$\pm$0.009 \\
SRA E, K=2$\to$4 (Cos.) & 12.52$\pm$0.02 & \textbf{12.89} & 0.219$\pm$0.003 & 0.326$\pm$0.002 \\
SRA B, K=1$\to$4 (Cos.) & 12.57$\pm$0.03 & 12.96 & 0.232$\pm$0.004 & 0.362$\pm$0.001 \\
StdMoE E, K=2$\to$4 (Lin.) & 12.57$\pm$0.02 & 13.18 & 0.240$\pm$0.007 & 0.325$\pm$0.005 \\
\hline
\end{tabular}
\end{table}

IC and EP are comparable across routing types. With full 3-seed validation across all four configurations, the IC gap between routing types is 0.018 for the B schedule and 0.021 for the E schedule (StdMoE slightly higher in both), while EP is essentially identical (0.001 for both B and E schedules). Notably, cosine routing exhibits lower cross-seed variance (IC $\pm$0.003--0.004 vs $\pm$0.007--0.011; EP $\pm$0.001--0.002 vs $\pm$0.005--0.009), suggesting cosine routing produces more stable specialization patterns regardless of the $k$ schedule. The E variants (K=2$\to$4) show lower IC and EP than B variants (K=1$\to$4) across both routing types, indicating that the initial top-1 phase may contribute to stronger specialization. On the test set (seed=19), cosine routing outperforms linear routing for both K schedules (SRA~E: 12.89 vs StdMoE~E: 13.18; SRA~B: 12.96 vs StdMoE~B: 13.01), narrowing or reversing the validation-set gap. This confirms that the training recipe---not the routing mechanism---drives semantic specialization. Both cosine and linear routing produce experts with similar quality of semantic clustering when trained with the same bandpass loss and progressive $k$ schedule. Section~\ref{sec:routing_eval} extends this analysis with routing-space metrics and LLM-as-a-judge evaluation.

\subsection{Extended Specialization Metrics}
\label{sec:extended_metrics}
Following the analysis framework of OLMoE \cite{Muennighoff2024}, we computed three additional specialization metrics across all five 256-expert models, summarized in Table~\ref{tab:extended_metrics}: \emph{router saturation} (normalized entropy of the per-token routing probability distribution), \emph{vocabulary specialization} (normalized entropy of each expert's token distribution---lower means more specialized), and \emph{expert co-activation} (pairwise frequency of experts selected together for the same token). Metrics are computed over the full validation set at $k$=4.

\begin{table}[h]
\caption{\textbf{Extended specialization metrics (cross-layer averages, 256 experts, $k$=4). Router Ent. = normalized routing entropy ($\uparrow$=more uniform); Vocab Ent. = normalized per-expert vocabulary entropy ($\downarrow$=more specialized). $\dagger$Dead = absolute count of dead experts summed across all 4 layers (seed=19 only); Table~\ref{tab:routing_comparison} reports percentage range across 3 seeds.}}
\label{tab:extended_metrics}
\centering
\begin{tabular}{|l|c|c|c|c|}
\hline
Model & Routing & Router Ent. & Vocab Ent. & Dead\textsuperscript{$\dagger$} \\
\hline
SRA B (K=1$\to$4) & Cosine & 0.865 & \textbf{0.330} & 1 \\
SRA E (K=2$\to$4) & Cosine & 0.917 & 0.384 & 0 \\
StdMoE B (K=1$\to$4) & Linear & 0.705 & 0.366 & 1 \\
StdMoE E (K=2$\to$4) & Linear & 0.851 & 0.416 & 0 \\
Vanilla StdMoE & Linear & 0.802 & 0.380 & 0 \\
\hline
\end{tabular}
\end{table}

Two patterns emerge. First, cosine routing provides \emph{more stable router saturation across layers}: in the matched K=1$\rightarrow$4 pair, SRA~B's per-layer normalized entropy ranges from 0.82--0.93 (range 0.11), while StdMoE~B varies from 0.47--0.93 (range 0.46). StdMoE~B Layer~2 exhibits highly peaked routing (entropy 0.47), indicating near-deterministic expert selection for most tokens---a pattern not observed in cosine routing, where the bounded cosine similarity range and temperature scaling naturally prevent extreme logit magnitudes.

Second, cosine routing yields \emph{tighter vocabulary specialization}: SRA~B's experts have normalized vocabulary entropy of 0.330 versus StdMoE~B's 0.366 (10\% more specialized). This pattern holds across both K-schedules (SRA~E: 0.384 vs StdMoE~E: 0.416, 8\% more specialized). Co-activation is uniformly low across all models ($\approx$0.0002), indicating that expert selection is largely independent regardless of routing type.

These metrics complement the IC/EP analysis: while emergent semantic specialization (IC/EP) is comparable between routing types, cosine routing produces structurally more regular routing patterns and more concentrated expert vocabularies---properties that enhance the practical value of routing-level inspection. Detailed visualization of these patterns is provided in the following sections (Sections~\ref{sec:utilization}--\ref{sec:overhead}).

\subsection{Expert Utilization}
\label{sec:utilization}
\begin{figure}[t]
\centering
\includegraphics[width=0.48\textwidth]{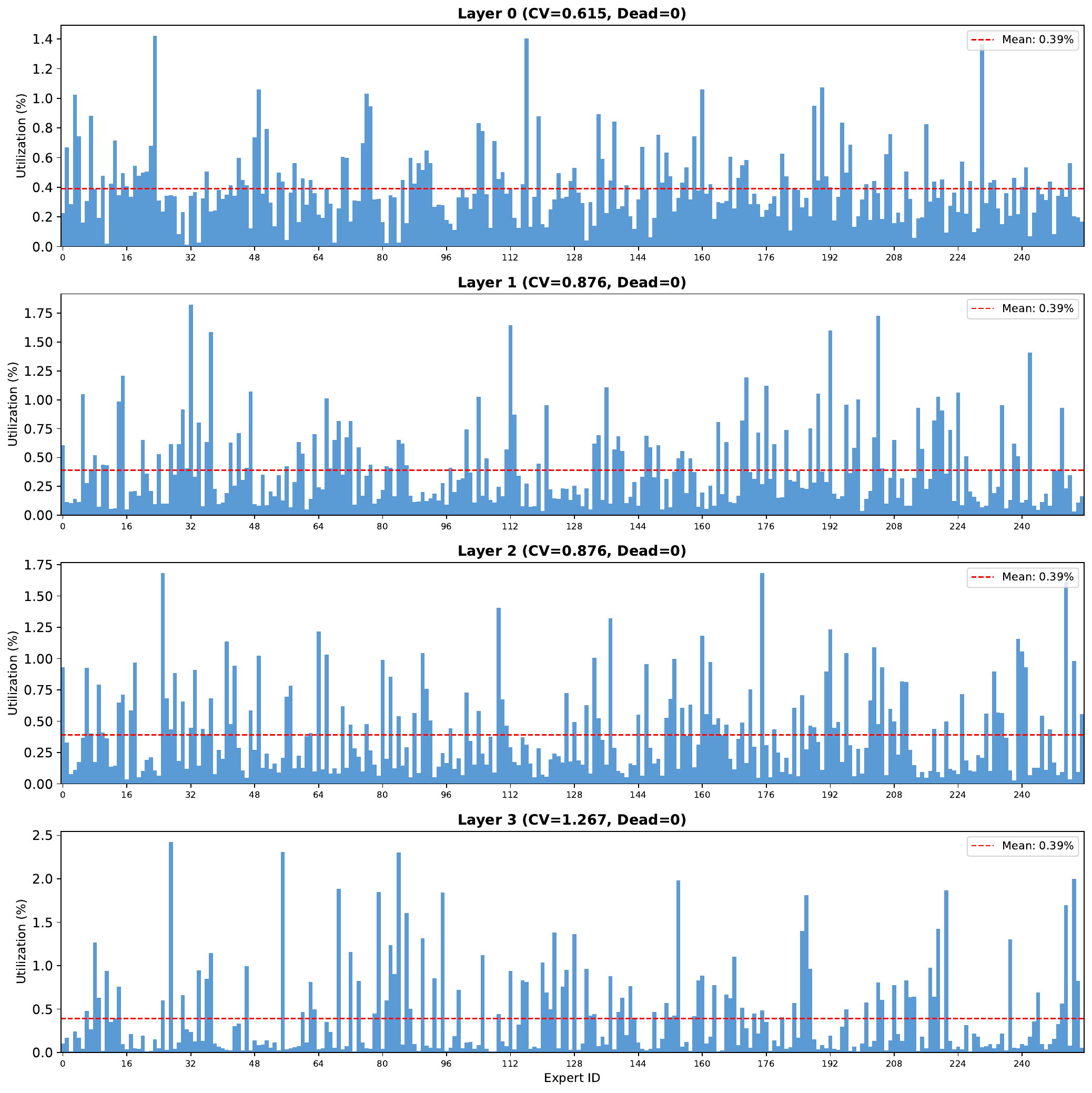}\hfill
\includegraphics[width=0.48\textwidth]{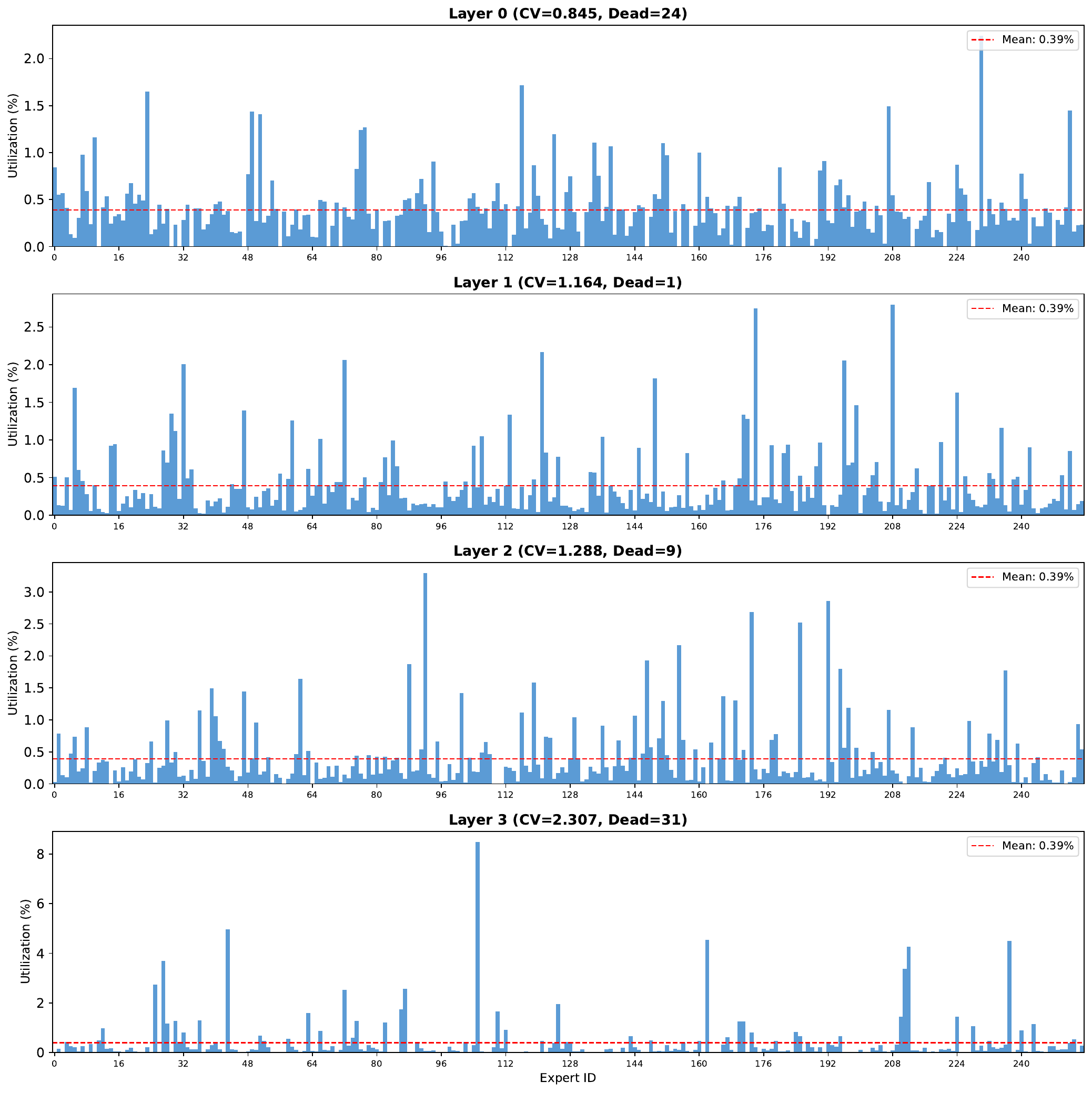}
\caption{\textbf{Expert utilization for SRA-E (256 experts, K=2$\to$4, seed=42).} Left: K=4 evaluation---zero dead experts across all layers. Right: K=1 evaluation (same model)---some experts are inactive at top-1 because SRA-E never trained with K=1; these experts contribute only in combinations.\label{fig:expert_utilization}}
\end{figure}

Figure~\ref{fig:expert_utilization} shows expert utilization across all four layers of the final SRA model.

The progressive routing strategy is essential for expert health in our experiments. During the initial top-1 phase (epochs 1--2), each token activates only one expert, allowing all experts to establish baseline specialization before competing in top-$k$ combinations. The ablation results (Table~\ref{tab:k_schedule}) confirm this: all schedules reaching K=4 (Runs A, B, D--G) achieve $\leq$6\% dead experts, while the K=2$\rightarrow$4 schedule (Run E) reaches 0 dead experts out of 1,024. Combined with the bandpass loss, progressive routing prevents the 30--32\% dead expert rates observed under earlier configurations (CV$^2$ at 128-expert scale; early bandpass corridor at 256-expert scale, Table~\ref{tab:main_results}).

Figure~\ref{fig:expert_utilization} (right) shows expert utilization when the same trained model is evaluated with $k$=1 (i.e., only the highest-scoring expert per token). Since SRA-E uses K=2$\rightarrow$4 (never training at K=1), some experts are inactive under top-1 selection---these experts contribute only in combinations. By contrast, B-schedule models (K=1$\rightarrow$4) show broader top-1 coverage because the initial K=1 phase forces each expert to establish standalone utility.

\subsection{Semantic Anchor Visualization}
\begin{figure}[t]
\centering
\includegraphics[width=\textwidth]{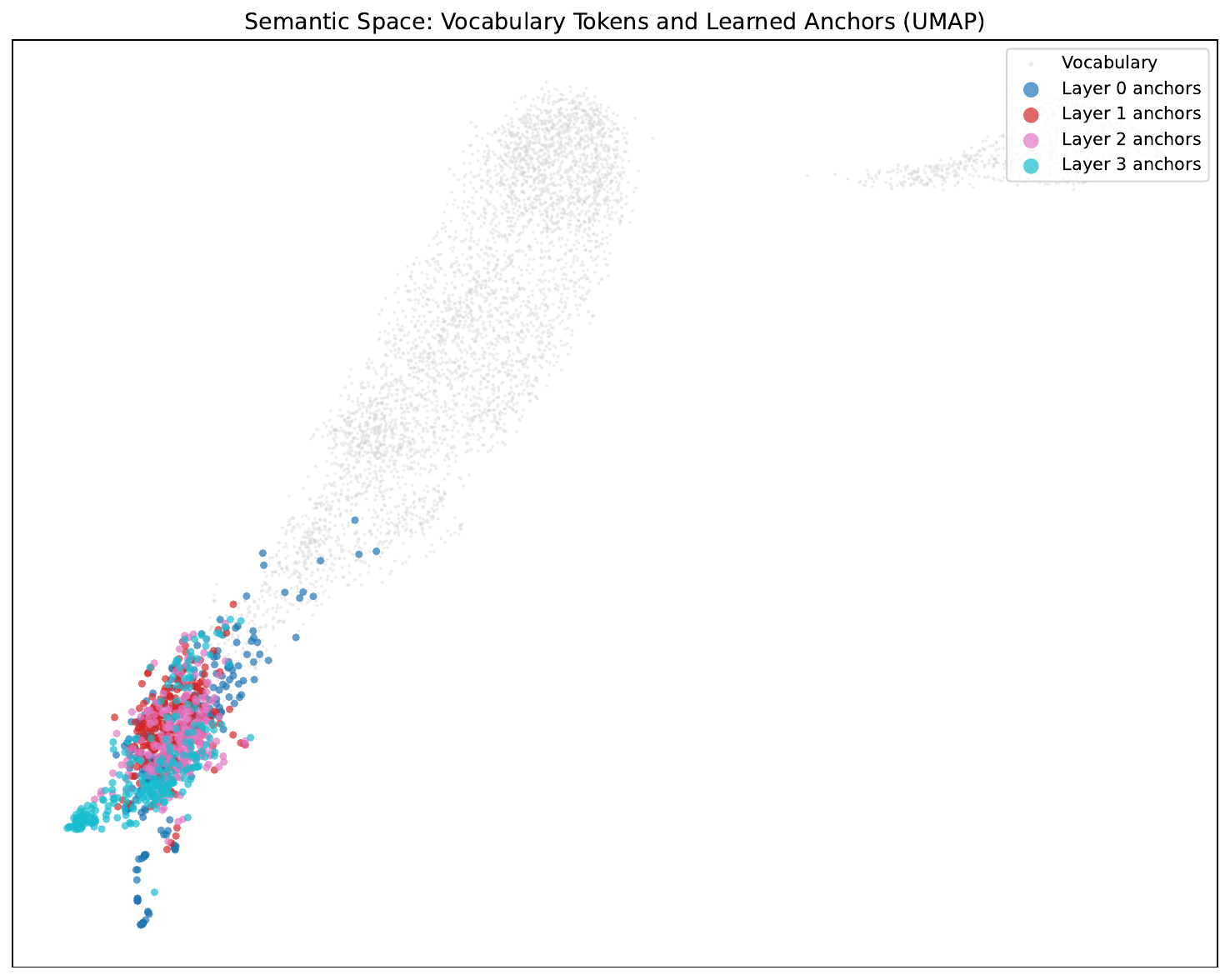}
\caption{\textbf{UMAP projection of the semantic space (SRA-E, 256 experts) showing vocabulary tokens (gray) and learned anchor positions by layer.}\label{fig:tsne_projection}}
\end{figure}

Figure~\ref{fig:tsne_projection} presents the UMAP projection of the learned semantic space.

\begin{figure}[t]
\centering
\includegraphics[width=\textwidth]{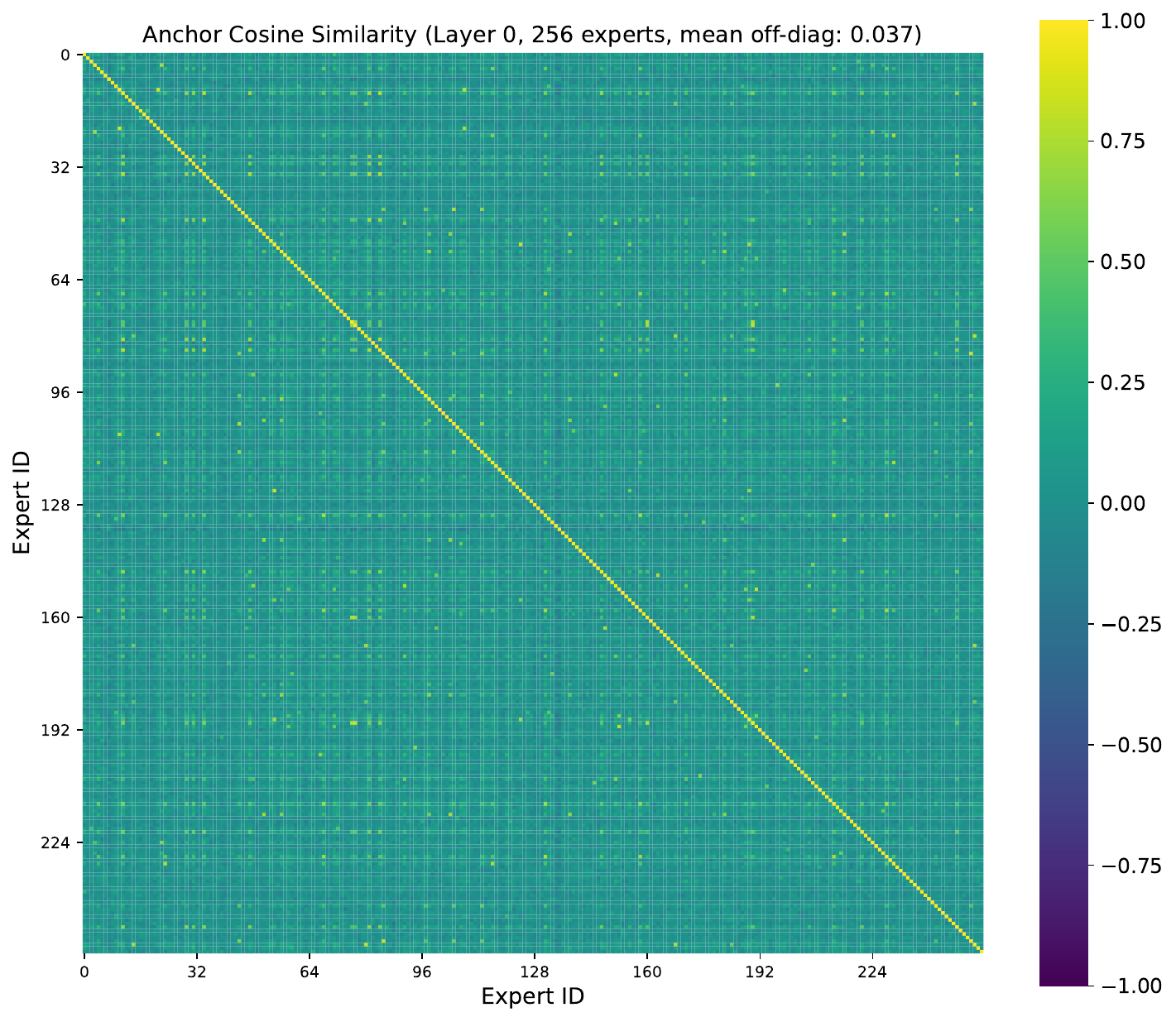}
\caption{\textbf{Cosine similarity heatmap between semantic anchors (Layer 0, 256 experts). Low off-diagonal values indicate effective anchor separation.}\label{fig:cosine_similarity_heatmap}}
\end{figure}

The visualization reveals hierarchical organization: early layer anchors cluster near high-frequency tokens (functional/syntactic roles), while later layers exhibit greater dispersion (diverse semantic categories). Figure~\ref{fig:cosine_similarity_heatmap} shows the anchor similarity matrix for Layer 0, with low off-diagonal values (mean pairwise similarity: 0.037) indicating successful anchor dispersion despite the higher expert count.

\subsection{Routing Overhead}
\label{sec:overhead}
Table~\ref{tab:overhead} compares the computational cost of cosine-similarity routing versus learned linear routing.

\begin{table}[h]
\caption{\textbf{Routing overhead comparison per token per layer (N=256 experts, D=512).}}
\label{tab:overhead}
\centering
\begin{tabular}{|l|c|c|}
\hline
Operation & Cosine Router & Linear Router \\
\hline
Routing FLOPs & $3ND + N$ & $ND$ \\
 & (norm + dot + scale) & (matmul) \\
Routing params & $ND$ (anchors) & $ND$ (gate) \\
Relative cost & $\sim$3$\times$ & 1$\times$ \\
\hline
\end{tabular}
\end{table}

Cosine routing requires approximately 3$\times$ the routing FLOPs due to L2 normalization of both the token and anchor vectors before the dot product. However, routing computation constitutes a small fraction of total forward-pass cost (dominated by expert FFN computation and attention): with $N$=256, $D$=512, and K=4 experts of $D_{ff}$=512, at the 256-expert scale with $D_{ff}$=512, routing accounts for a small fraction of per-layer FLOPs. At the 128-expert scale ($D_{ff}$=1024), training throughput is comparable between cosine and linear routing ($\sim$3.0 steps/s). At the 256-expert scale ($D_{ff}$=256), measured training throughput on a single RTX 3090 is $\sim$1.0--1.9 steps/s for cosine routing and $\sim$1.8--1.9 steps/s for linear routing, corresponding to $\sim$32k--62k tokens/s (batch size 128, sequence length 256). We note that the routing overhead ratio decreases as expert capacity grows: since routing cost scales as $O(ND)$ while expert FFN cost scales as $O(K \cdot D \cdot D_{ff})$, increasing $D_{ff}$ (as is typical at larger scales) makes the routing fraction negligible. For a 10B-parameter MoE with $D$=4096, $N$=256, and $D_{ff}$=8192, cosine routing would constitute $<$0.5\% of per-layer FLOPs.

\subsection{Interpretability Case Study}
To illustrate routing interpretability, we examined routing decisions for the sentence: ``The film was released in December 1995 and received positive reviews.'' Table~\ref{tab:case_study} details the top-2 expert assignments in Layer 0.

\begin{table}[t]
\caption{\textbf{Routing decisions for an example sentence (Layer 0, SRA-E, seed=42). Interpretations derived from each expert's top tokens.}}
\label{tab:case_study}
\centering
\begin{tabular}{|l|c|c|c|c|c|}
\hline
Token & E1 & W1 & E2 & W2 & Interpretation (E1 + E2 Top Tokens) \\
\hline
The & $E_{224}$ & 0.77 & $E_{4}$ & 0.23 & (The, In, This) + (the, and, in) \\
film & $E_{136}$ & 0.95 & $E_{0}$ & 0.05 & (film, Qu, movie) + (\textquotesingle, \textquotesingle, Were) \\
was & $E_{94}$ & 0.72 & $E_{197}$ & 0.28 & (was, had, would) + (was, had, became) \\
released & $E_{27}$ & 0.89 & $E_{94}$ & 0.11 & (recorded, released, built) + (was, had, would) \\
in & $E_{10}$ & 0.50 & $E_{21}$ & 0.50 & (in, on, during) + (in, within, during) \\
December & $E_{240}$ & 0.74 & $E_{196}$ & 0.26 & (August, September, July) + (La, Er, par) \\
1995 & $E_{75}$ & 0.56 & $E_{181}$ & 0.44 & (Tri, cost, red) + (national, intl., Pol) \\
and & $E_{252}$ & 0.69 & $E_{105}$ & 0.31 & (and, but, or) + (and, with, but) \\
received & $E_{62}$ & 0.76 & $E_{197}$ & 0.24 & (Dr, considered, season) + (was, had, became) \\
positive & $E_{41}$ & 0.81 & $E_{240}$ & 0.19 & (is, Bell, Mah) + (August, Sept., July) \\
reviews & $E_{255}$ & 0.72 & $E_{192}$ & 0.28 & (--, und, \%) + (Bill, French, music) \\
. & $E_{104}$ & 0.80 & $E_{138}$ & 0.20 & (., ``., ;) + (., .'', ``.) \\
\hline
\end{tabular}
\end{table}

Routing decisions align with expert specializations: ``film'' routes to $E_{136}$ (film/movie tokens, weight 0.95), ``released'' routes to $E_{27}$ (recorded/released/built) and ``December'' routes to $E_{240}$ (month names). Auxiliary verbs share experts: ``was'' and ``received'' both route to $E_{94}$/$E_{197}$ (was/had/would/became). Conjunctions and prepositions receive clear assignments (``and'' $\rightarrow$ $E_{252}$: and/but/or; ``in'' splits evenly between two preposition experts). Some Layer 0 experts reflect orthographic rather than semantic patterns, consistent with the finding that 44--54\% of specialization is syntactic (Section~\ref{sec:routing_eval}). In cosine routing, these assignments are directly traceable to the similarity between the token's representation and the expert's anchor vector---a property not available with linear gates, where understanding \emph{why} a token was routed to an expert requires additional analysis.

\subsection{Routing-Space Interpretability Evaluation}
\label{sec:routing_eval}
To complement the embedding-space IC/EP analysis with evidence from actual routing decisions, we evaluate expert specialization using three approaches: information-theoretic metrics, LLM-as-a-judge evaluation \cite{Zheng2023}, and word-level routing coherence. Information-theoretic metrics and routing coherence are computed over the full validation set at $k$=4 and averaged across 3 seeds (19, 42, 137); LLM-as-a-judge evaluation uses seed 137 with full expert coverage (all experts per layer), supplemented by seeds 19 and 42 on stratified samples.

\subsubsection{Information-Theoretic Metrics}
We compute Normalized Mutual Information (NMI) between tokens and expert assignments as a global monosemanticity measure, and mean pairwise Jensen-Shannon (JS) divergence between expert token distributions to quantify expert distinctness. Table~\ref{tab:routing_space} confirms the IC/EP finding: global routing-space monosemanticity is comparable between routing types (NMI $\Delta$<0.01). SRA experts show slightly higher mean JS divergence (0.808$\pm$0.001 vs 0.805$\pm$0.004 across seeds), though the difference is small and overlaps within seed variance, consistent with the lower vocabulary entropy in Table~\ref{tab:extended_metrics}.

\begin{table}[h]
\caption{\textbf{Routing-space metrics (cross-layer mean, 3 seeds). NMI($\uparrow$) = more structured routing; JS Div($\uparrow$) = more distinct expert token distributions.}}
\label{tab:routing_space}
\centering
\begin{tabular}{|l|c|c|}
\hline
Routing & NMI & JS Div \\
\hline
Cosine (SRA B) & 0.515$\pm$0.006 & 0.808$\pm$0.001 \\
Linear (StdMoE B) & 0.512$\pm$0.006 & 0.805$\pm$0.004 \\
\hline
\end{tabular}
\end{table}

\subsubsection{LLM-as-a-Judge Expert Purity}
We evaluated expert semantic coherence using an LLM-as-a-judge protocol. For seed 137, we evaluated \emph{every} expert across all 4 layers of both models (2,041 of 2,048 experts---a full census; 7 dead experts with no routed tokens were excluded). Each expert's top-10 tokens ranked by pointwise mutual information (PMI) were presented to three independent LLM evaluators---Gemini~3.1~Pro (Google), GPT-5.2-Codex (OpenAI), and Claude~Opus~4.6 (Anthropic)---who scored semantic coherence on a 1--10 scale following a standardized rubric with calibration anchors. Evaluation was blinded: expert IDs were anonymized, row order randomized, and IC/EP metrics hidden. Seeds 19 and 42 were additionally evaluated using stratified-random samples of 100 experts per layer (25 per utilization quartile) with top-50 PMI tokens, totaling 1,600 experts across those two seeds.

\begin{table}[h]
\caption{\textbf{LLM-as-a-judge expert purity (seed 137, all 2,041 experts---full census). Top-10 PMI tokens per expert. $\Delta$ = SRA $-$ StdMoE mean score. Repr.~$r$ = Pearson $r$ between two independent scoring rounds on identical data.}}
\label{tab:llm_judge}
\centering
\begin{tabular}{|l|c|c|c|c|}
\hline
Evaluator & SRA B & StdMoE B & $\Delta$ & Repr.~$r$ \\
\hline
Gemini 3.1 Pro & 7.32 & 7.20 & +0.12 & 0.843 \\
GPT-5.2-Codex & 6.86 & 6.72 & +0.14 & \textbf{0.888} \\
Claude Opus 4.6 & 5.95 & 5.76 & +0.18 & \textbf{0.933} \\
\hline
\end{tabular}
\end{table}

All three evaluators assign marginally higher mean scores to SRA (Table~\ref{tab:llm_judge}), with consistent $\Delta$=+0.12 to +0.18 and no verdict reversals between rounds. Reproducibility is high (Pearson $r$=0.84--0.93), validating the protocol beyond one-shot LLM scoring. However, Mann-Whitney U tests with Holm-Bonferroni correction show that pooled differences are negligible for all evaluators (Cliff's $\delta$=0.03--0.04, $p>$0.10). One evaluator (Opus) shows significant per-layer effects, but these alternate in sign across layers and cancel in the pooled analysis; the other two evaluators show no per-layer effects, suggesting evaluator-specific noise rather than a systematic architectural signal. In the supplementary evaluation on seeds 19 and 42 using top-50 PMI tokens with 100-expert stratified samples, no evaluator consensus emerged ($\Delta$ ranged from $-$0.38 to +0.25), likely because with 50 tokens stopwords dilute the signal despite PMI ranking.

IC (embedding-space internal cohesion) has \emph{near-zero correlation} with LLM-judged purity (Spearman $r$=$-$0.05 to $-$0.14 across evaluators). With the full census ($n$=2,041), these weak correlations reach statistical significance ($p<$0.02), but the effect sizes are negligible---IC explains less than 2\% of variance in purity scores. The increase in significance from the supplementary top-50 evaluation ($n$=1,600, $p>$0.05) reflects increased statistical power, not a change in effect magnitude. Among automated metrics, vocabulary entropy shows the strongest relationship with LLM-judged purity ($r$=$-$0.13 to $-$0.35, all $p<$0.001), explaining up to 12\% of variance---experts with more concentrated token distributions are judged as more monosemantic. This substantially outperforms IC as a predictor of expert coherence.

All three evaluators converge on a qualitative finding: \textbf{44--54\% of experts specialize syntactically} (punctuation, suffixes, function words), 18--37\% semantically (topical categories), and 18--32\% mixed, with no significant difference in type distribution between routing types ($\chi^2$ test, $p>$0.06 for all evaluators). This suggests that high monosemanticity scores reflect substantial syntactic structure in expert routing, qualifying claims about semantic specialization in MoE models.

\subsubsection{Word-Level Routing Coherence}
To test whether cosine routing maintains more consistent subtoken grouping, we measured routing coherence for 100 multi-token words (BPE-split into 2--5 subtokens) across 5 categories, each embedded in 10 carrier sentences with varied syntax and target position. Coherence is the fraction of subtokens assigned to the same top-1 expert (random baseline: $1/256 \approx 0.004$). Ten nonce words served as controls.

\begin{table}[h]
\caption{\textbf{Word-level routing coherence by layer (3 seeds, 100 words $\times$ 10 templates). Random baseline = 0.004. $^{*}$Permutation test $p<$0.001.}}
\label{tab:routing_coherence}
\centering
\begin{tabular}{|l|c|c|}
\hline
Layer & SRA B & StdMoE B \\
\hline
0 & 0.447 & \textbf{0.451} \\
1 & \textbf{0.433} & 0.425 \\
2 & \textbf{0.464}$^{*}$ & 0.454 \\
3 & \textbf{0.469}$^{*}$ & 0.443 \\
\hline
Mean & \textbf{0.453} & 0.443 \\
\hline
\end{tabular}
\end{table}

Both models route subtokens far above the random baseline (Table~\ref{tab:routing_coherence}), confirming meaningful word-level routing structure. Cosine routing shows a consistent advantage in deeper layers (2--3); permutation tests (10,000 iterations) confirm this difference is significant ($p<$0.001 for both layers). The effect is strongest on technical vocabulary (Cliff's $\delta$=+0.06) and 5-subtoken words ($\delta$=+0.22), and weakest on nonce words---consistent with the hypothesis that anchor alignment benefits semantically coherent tokens.

\subsection{Robustness}
\label{sec:robustness}
To assess sensitivity to random initialization, we train all four routing configurations (Table~\ref{tab:ic_ep}) across 3 random seeds (19, 42, 137)---a total of 12 training runs. For the B schedule (K=1$\to$4), linear routing achieves 12.45$\pm$0.03 PPL and cosine routing 12.57$\pm$0.03. A paired $t$-test on the three seed pairs yields $t$=4.29, $p$=0.050, confirming that the 0.12 PPL gap is consistent in direction across all seeds, though borderline significant given $n$=3. The E schedule (K=2$\to$4) shows a reversed pattern: cosine routing (12.52$\pm$0.02) outperforms linear (12.57$\pm$0.02), suggesting the initial top-1 phase interacts differently with the two routing mechanisms. All four configurations exhibit low variance ($\pm$0.02--0.03), demonstrating training stability. Dead expert counts are stable: 0--5\% for linear and 0--6\% for cosine across all seeds, with E variants showing 0\% dead experts in all runs. Our broader ablation study spanning 17 configurations provides additional trend-based evidence: (1) progressive routing reduces dead experts compared to fixed-$k$ training (7 configurations); (2) bandpass loss outperforms CV$^2$ on expert utilization (5 loss configurations); (3) higher $k$ improves perplexity (3 matched K-schedule pairs); (4) temperature $\tau$=10.0 is near-optimal (learned temperatures converge to $\sim$9.7 regardless of initialization from 5.0, 10.0, or 20.0); and (5) IC/EP differences between cosine and linear routing are small in magnitude across all matched pairs (IC gap 0.018--0.021, EP gap 0.001 with 3-seed validation across both schedules). The balance weight $\alpha$=0.4 was confirmed optimal via a fine-grained sweep (0.35, 0.40, 0.45, 0.50), with all values within 0.06 perplexity of each other.

\subsection{Cross-Dataset Validation}
\label{sec:cross_dataset}
To assess generalization beyond WikiText-103, we trained matched cosine and linear routing models on OpenWebText \cite{Gokaslan2019}, a corpus of web text with different domain characteristics. We sampled 106K documents ($\approx$408K training chunks at sequence length 256, matching WikiText-103's 405K chunks), trained a new BPE tokenizer (32K vocabulary), and used identical hyperparameters to the E-schedule WikiText-103 experiments (256 experts, $D_{ff}$=256, K=2$\rightarrow$4, bandpass loss, $\alpha$=0.4, seed=19).

\begin{table}[h]
\caption{\textbf{Cross-dataset validation: WikiText-103 vs OpenWebText (E schedule, K=2$\to$4, 256 experts, seed=19). WT103 reports mean$\pm$std across 3 seeds; OWT is single seed.}}
\label{tab:cross_dataset}
\centering
\begin{tabular}{|l|l|c|c|c|c|}
\hline
Dataset & Routing & Val PPL & Dead & IC & EP \\
\hline
WT103 & Cosine (SRA E) & 12.52$\pm$0.02 & 0\% & 0.219 & 0.326 \\
WT103 & Linear (StdMoE E) & 12.57$\pm$0.02 & 0\% & 0.240 & 0.325 \\
\hline
OWT & Cosine (SRA E) & \textbf{44.88} & 0\% & 0.212 & 0.327 \\
OWT & Linear (StdMoE E) & 45.44 & 0\% & 0.238 & 0.332 \\
\hline
\end{tabular}
\end{table}

Three findings transfer from WikiText-103 to OpenWebText (Table~\ref{tab:cross_dataset}). First, cosine and linear routing achieve comparable perplexity on the new domain (44.88 vs 45.44), with cosine routing slightly ahead---the same pattern as the E schedule on WikiText-103. Second, the bandpass loss with K=2$\rightarrow$4 produces zero dead experts across all 1,024 experts in both models, confirming that the training recipe generalizes. Third, the IC/EP relationship between routing types is preserved: linear routing shows marginally higher IC (0.238 vs 0.212) while EP is comparable (0.332 vs 0.327), consistent with the WikiText-103 finding that the training recipe---not the routing function---drives specialization quality. The higher absolute perplexity on OpenWebText reflects the greater diversity and noise of web text compared to curated Wikipedia articles, not a degradation of the architecture.

\section{Discussion}

\subsection{Cosine Routing as an Inspectable Mechanism}
The SRA architecture provides routing-level interpretability through a direct and measurable mechanism: every routing decision is determined by the cosine similarity between the token representation and the expert's semantic anchor. Unlike learned linear gates, where the routing decision is a function of an opaque weight matrix, cosine routing produces scores that have a geometric interpretation---the angular distance between the token and anchor in the shared representation space.

Our matched routing comparison (Section~\ref{sec:routing_comparison}), now validated across all four configurations with 3 seeds each (12 runs total), reveals an important finding: cosine routing does not produce inherently better semantic specialization than linear routing when both are trained with the same recipe. The IC and EP metrics are comparable (IC gap 0.018--0.021, EP gap 0.001 across both $k$ schedules), with cosine routing exhibiting lower cross-seed variance (IC $\pm$0.003--0.004 vs $\pm$0.007--0.011; EP $\pm$0.001--0.002 vs $\pm$0.005--0.009). For the primary B schedule, linear routing achieves marginally better perplexity (12.45$\pm$0.03 vs 12.57$\pm$0.03), while the E schedule reverses this ordering (cosine 12.52$\pm$0.02 vs linear 12.57$\pm$0.02). This means the value of cosine routing lies not in emergent specialization quality but in its \emph{built-in inspectability}---one can examine the anchor vectors directly and compute their similarity to any token, providing immediate insight into routing behavior without post-hoc probing tools.

However, the extended specialization metrics (Section~\ref{sec:extended_metrics}) reveal structural differences that the IC/EP analysis does not capture. Cosine routing maintains more stable router saturation across layers (entropy range 0.11 vs 0.46 for linear) and produces tighter per-expert vocabulary distributions (10\% lower vocabulary entropy). These properties are consistent with the bounded nature of cosine similarity, which limits logit magnitudes and may prevent the layer-specific saturation collapse observed in linear routing. While these structural differences do not translate to better emergent semantic clustering (as measured by IC/EP), they make the routing patterns more regular and thus more amenable to systematic analysis.

The routing-space evaluation (Section~\ref{sec:routing_eval}) provides broadly consistent evidence from three independent approaches. Information-theoretic metrics (NMI, JS divergence) computed from actual routing decisions confirm IC/EP near-parity, while a full-census LLM-as-a-judge evaluation of all 2,041 experts found no statistically significant difference in expert purity between routing types (all three evaluators yielded pooled Cliff's $\delta<$0.05, $p>$0.10). The consistent direction of all three evaluators marginally favoring SRA ($\Delta$=+0.12 to +0.18) suggests a possible weak advantage that would require substantially larger expert populations to confirm. Word-level routing coherence reveals that cosine routing provides significantly better subtoken consistency in deeper layers (permutation test $p<$0.001), a structural advantage that global metrics do not capture. Among automated metrics, vocabulary entropy outperforms IC as a predictor of LLM-judged purity ($|r|$ up to 0.35 vs $<$0.14), suggesting that vocabulary concentration is a more reliable indicator of expert specialization than information-theoretic clustering metrics. All evaluators agree that 44--54\% of expert specialization is syntactic rather than semantic, indicating that monosemanticity metrics may overstate the degree of semantic organization in both routing types.

\subsection{Bandpass Loss as a Transferable Contribution}
The bandpass routing loss is a general-purpose contribution that improves expert utilization across routing types. The key insight is that the ceiling parameter (anti-monopoly constraint) matters more than the floor (anti-death constraint): monotonic improvement in perplexity was observed as the ceiling tightened toward the uniform rate. Combined with PAD masking, the bandpass loss reduced dead experts from 30--32\% (under earlier configurations) to 0--6\%, while simultaneously improving perplexity. Importantly, the bandpass loss with PAD masking improved StdMoE performance from 12.76 (CV$^2$, K=1$\rightarrow$2) to 12.45$\pm$0.03 (bandpass, K=1$\rightarrow$4, across 3 seeds), demonstrating that the training recipe innovations transfer directly to standard MoE architectures.

\subsection{Deployment Considerations}
\label{sec:deployment}
SRA's total parameter count ($\approx$559M) is substantially larger than the dense baseline ($\approx$29M), even though active parameters per token are matched. This implies a $\sim$19$\times$ memory overhead for storing expert weights, which is the standard trade-off for all MoE architectures. At the scale tested (559M total, 29M active), this is manageable on a single GPU. For larger deployments, expert parallelism across multiple devices would be necessary, following the same strategies used by other MoE systems \cite{Fedus2022, Lepikhin2021}.

Cosine routing adds modest routing overhead ($\sim$3$\times$ routing FLOPs; Section~\ref{sec:overhead}), with measured throughput of 1.0--1.9 vs 1.8--1.9 steps/s depending on configuration. The inference-time K-sweep (Section~\ref{sec:k_sweep}) provides a practical deployment benefit: setting K=5 at inference yields a free $\sim$0.1--0.2 perplexity improvement over the best K=4 validation PPL from training without retraining, offering a compute--quality trade-off that can be tuned at deployment time.

\subsection{Future Work}
Future directions include: (1) scaling to larger expert capacity ($D_{ff}$=512 and beyond) with the optimized bandpass corridor and multi-seed validation; (2) scaling to billion-parameter models to verify that the routing-level interpretability properties and bandpass loss benefits hold at larger scale; (3) applying the bandpass loss to other MoE architectures (e.g., Mixtral, DeepSeekMoE) as a drop-in replacement for standard load balancing; (4) evaluating on downstream tasks (question answering, dialogue, code) to assess whether routing-level interpretability provides actionable insights beyond language modeling; and (5) developing adaptive routing strategies that dynamically determine the number of active experts based on input complexity.

\subsection{Broader Impacts}
The routing-level interpretability provided by SRA opens potential research directions for model analysis. By examining routing patterns and anchor--token similarity, researchers could investigate whether specific token categories are systematically processed by particular experts, potentially providing insights into model behavior. However, we emphasize that these are research directions rather than established capabilities---using routing patterns for applications such as bias detection or content control would require dedicated experiments and careful validation beyond the scope of this work. The bandpass routing loss, as a training technique that improves expert utilization, has direct practical value for any MoE deployment regardless of interpretability goals.

\subsection{Contributions}
The contributions of this work are: (1) cosine-similarity routing with semantic anchors, providing an inspectable routing mechanism where every decision is grounded in measurable similarity scores; (2) the bandpass routing loss with PAD masking, a transferable technique that improves expert utilization from 68--70\% to 94--100\% across both cosine and linear routing; (3) a comprehensive interpretability evaluation spanning 17 configurations, including IC/EP, OLMoE-style extended metrics, routing-space information-theoretic analysis (NMI, JS divergence), LLM-as-a-judge expert scoring with three independent evaluators, and word-level routing coherence, establishing that the training recipe drives specialization quality while cosine routing provides inherent inspectability, structurally more regular routing patterns, and better subtoken routing consistency in deeper layers; and (4) the finding that cosine routing produces more stable router saturation across layers and tighter per-expert vocabulary specialization than linear routing, even when emergent semantic clustering is comparable, with the novel observation that 44--54\% of expert specialization in both routing types is syntactic rather than semantic.

\subsection{Limitations}
The limitations of this work should be acknowledged. First, while cross-dataset validation on OpenWebText (Section~\ref{sec:cross_dataset}) confirms that the core findings generalize beyond WikiText-103, all experiments remain at a medium scale ($\sim$295--559M parameters) on language modeling tasks, and results may not directly generalize to larger models or downstream tasks such as question answering or dialogue. Second, an exploratory larger configuration (559M params, $D_{ff}$=512) achieved 12.20 PPL in a single-seed run but used an early bandpass corridor with 30\% dead experts; this result has not been multi-seed validated and should be viewed as a scaling data point rather than a validated claim. Third, the matched routing comparison shows that cosine routing does not improve semantic specialization over linear routing with the same recipe, and the perplexity ordering depends on the $k$ schedule (linear wins B: 12.45 vs 12.57; cosine wins E: 12.52 vs 12.57)---the interpretability advantage is in inspectability of the routing mechanism itself, not in perplexity or specialization quality. Fourth, 256-expert configurations exhibit high dead expert rates (30--32\%, CV$^2$ at 128-expert scale, early bandpass corridor at 256-expert scale) without the optimized bandpass loss, indicating that SRA's effectiveness depends on careful training procedure design. Fifth, the cosine-similarity computation introduces $\sim$3$\times$ routing overhead relative to linear gating; while this is modest at the current scale, it may become more significant at very large expert counts. Sixth, while semantic anchors provide inspectable routing, the internal computation within each expert remains opaque---SRA does not provide full mechanistic transparency. The LLM-as-a-judge evaluation (Section~\ref{sec:routing_eval}) relies on automated evaluators without human ground truth; while reproducibility testing on the full-census evaluation showed good self-consistency ($r$=0.84--0.93 between independent scoring rounds), evaluator calibration varies substantially (mean scores 5.8--7.3), and the pooled SRA--StdMoE differences do not reach statistical significance. Human evaluation would provide a more authoritative baseline. The pooled statistical analysis treats experts across layers as independent; within-model shared training dynamics may reduce effective sample sizes. Finally, the semantic anchors trained for one domain may not transfer to others without fine-tuning.

\section{Conclusion}
This work introduced the Semantic Resonance Architecture, which routes tokens to experts via cosine similarity with learnable semantic anchors. A controlled multi-seed comparison on WikiText-103 (3 seeds $\times$ 4 configurations = 12 runs, 256 experts, $D_{ff}$=256) demonstrates that cosine routing achieves competitive perplexity with standard linear routing (B schedule: 12.57$\pm$0.03 vs 12.45$\pm$0.03; E schedule: 12.52$\pm$0.02 vs 12.57$\pm$0.02) and comparable global specialization metrics when trained with the same recipe. The optimized E-schedule configuration achieves 12.52$\pm$0.02 PPL with zero dead experts across all seeds, compared to a dense baseline (14.13) and standard MoE (12.57$\pm$0.02), demonstrating effective sparse expert utilization at matched active parameter constraints ($\approx$29M). Cosine routing provides significantly better word-level routing coherence in deeper layers ($p<$0.001, permutation test), suggesting a more structurally consistent mapping of subtoken units to experts. Embedding-space metrics (IC) show near-zero correlation with LLM-judged expert purity ($|r|<$0.14, $<$2\% variance explained), indicating these approaches capture different facets of specialization. Notably, 44--54\% of expert specialization in both routing types is syntactic rather than semantic, qualifying what monosemanticity means in practice. Cosine routing's primary advantage is its inherent inspectability---every routing decision can be traced to measurable anchor--token similarity---rather than improved specialization quality.

We introduced the bandpass routing loss, a floor-and-ceiling corridor constraint that reduces dead experts from 30--32\% (under earlier configurations) to 0--6\% when fully optimized, and transfers effectively to both routing types. Combined with PAD masking and progressive routing, this training recipe enables effective MoE training with up to 256 experts per layer. An inference-time $k$-sweep revealed that K=5 is optimal or tied-optimal across all models trained with K=4, providing a free deployment-time improvement.

Cross-dataset validation on OpenWebText confirms that these findings generalize: cosine and linear routing achieve comparable perplexity (44.88 vs 45.44), the bandpass loss eliminates dead experts, and IC/EP patterns are preserved on web-domain text. These results demonstrate that cosine-similarity routing provides an inspectable routing mechanism for MoE models with measurable subtoken consistency advantages, while the bandpass routing loss and associated training innovations represent transferable contributions applicable to the broader MoE literature. The finding that the majority of expert specialization is syntactic rather than semantic---observed consistently across routing types, seeds, datasets, and independent evaluators---suggests that monosemanticity metrics in MoE models warrant careful interpretation.

\section*{Acknowledgment}
We acknowledge the contributions of the PyTorch, HuggingFace (Accelerate), and Microsoft DeepSpeed teams to the open-source ecosystem, and wandb.ai for experiment tracking. The authors utilized Google Gemini 3.1 Pro for assistance in translating the manuscript into English and for grammatical proofreading throughout the text. Gemini 3.1 Pro, OpenAI GPT-5.2-Codex, and Anthropic Claude Opus 4.6 were used as independent evaluators in the LLM-as-a-judge expert purity assessment (Section~\ref{sec:routing_eval}). The scientific content, experimental design, and conclusions were generated solely by the authors.

\section*{Data and Code Availability}
Source code, training configurations, and experiment logs are available at \url{https://github.com/ITernovtsii/semantic-resonance}. Pre-trained checkpoints for all multi-seed and cross-dataset experiments are hosted at \url{https://huggingface.co/iternovtsii/sra-checkpoints}.

\section*{Changelog}

\textbf{v2} (current). Major revision expanding a single-seed proof-of-concept into a comprehensive evaluation:
\begin{itemize}
    \item \textbf{Bandpass routing loss} replaces the dispersion loss from v1. The new floor-and-ceiling corridor constraint reduces dead experts from 30--32\% to 0--6\% and, crucially, transfers to standard linear-gated MoE---improving StdMoE perplexity from 12.76 to 12.45$\pm$0.03.
    \item \textbf{Matched routing comparison} (3 seeds $\times$ 4 configurations) demonstrates that cosine and linear routing achieve comparable perplexity (12.57$\pm$0.03 vs 12.45$\pm$0.03 for K=1$\to$4). The training recipe---not the routing function---drives specialization quality; cosine routing's advantage is inherent inspectability.
    \item \textbf{Expanded experiments}: 17 configurations (up from 3), multi-seed validation, progressive top-$k$ schedules, inference-time $k$-sweep (free $\sim$0.1--0.2 PPL gain at $k$=5 over the best $k$=4 validation PPL), and cross-dataset validation on OpenWebText.
    \item \textbf{Quantitative interpretability}: Internal Cohesion / External Purity metrics, OLMoE-style extended analysis (router saturation, vocabulary specialization, co-activation), LLM-as-a-judge evaluation, and word-level subtoken coherence tests revealing 44--54\% syntactic specialization.
    \item \textbf{PAD masking} in routing statistics eliminates a systematic bias that inflated dead-expert counts.
    \item \textbf{Full-distribution softmax} before top-$k$ selection ensures gradient flow to all anchors even at $k$=1.
    \item \textbf{Learnable temperature} $\tau$ (initialized to 10.0) replaces fixed scaling.
    \item Claims toned down: v1 claimed superior specialization from cosine routing; v2 clarifies that the advantage is inspectability and structural consistency (more stable saturation, tighter vocabulary distributions), not emergent clustering quality.
\end{itemize}

\textbf{v1}. Initial preprint with cosine-similarity routing, dispersion loss, and single-seed evaluation on WikiText-103 (3 configurations).

\end{document}